\documentclass{article}
\usepackage{authblk}
\usepackage{amsfonts}

\usepackage{hyperref}
\usepackage[utf8]{inputenc}
\usepackage[textwidth=18cm]{geometry}
\usepackage{float, graphicx, amsmath, xcolor, caption, subcaption, hyperref}
\usepackage[sorting=none]{biblatex}
\bibliography{references} 

\title{Predictive Coding, Precision and Natural Gradients}
\author[1*]{Andre Ofner}
\author[1]{Raihan Kabir Ratul}
\author[1]{Suhita Ghosh}
\author[1]{Sebastian Stober}
\affil{Otto-von-Guericke University, Magdeburg, Germany \protect\\ ofner@ovgu.de}

\begin{document}

\maketitle

\begin{abstract}
There is an increasing convergence between biologically plausible computational models of inference and learning with local update rules and the global gradient-based optimization of neural network models employed in machine learning. One particularly exciting connection is the correspondence between the locally informed optimization in predictive coding networks and the error backpropagation algorithm that is used to train state-of-the-art deep artificial neural networks. Here we focus on the related, but still largely under-explored connection between precision weighting in predictive coding networks and the Natural Gradient Descent algorithm for deep neural networks. Precision-weighted predictive coding is an interesting candidate for scaling up uncertainty-aware optimization -- particularly for models with large parameter spaces -- due to its distributed nature of the optimization process and the underlying local approximation of the Fisher information metric, the adaptive learning rate that is central to Natural Gradient Descent. Here, we show that hierarchical predictive coding networks with learnable precision indeed are able to solve various supervised and unsupervised learning tasks with performance comparable to global backpropagation with natural gradients and outperform their classical gradient descent counterpart on tasks where high amounts of noise are embedded in data or label inputs. When applied to unsupervised auto-encoding of image inputs, the deterministic network produces hierarchically organized and disentangled embeddings, hinting at the close connections between predictive coding and hierarchical variational inference.
\end{abstract}

\section{Introduction}
Predictive coding is a theory of brain function originating from cognitive neuroscience. It casts the brain’s main function as the minimization of prediction errors that are generated with respect to a generative model of the world \cite{friston2009predictive, millidge2021predictivereview}. Predictive coding, jointly with related process theories that revolve around Bayesian approaches to brain function, such as active inference, is highly influential in its field of origin, cognitive and computational neuroscience \cite{friston2017active}. A core aspect of predictive coding models is their biological plausibility, which is backed up by a plethora of neurophysiological evidence. Predictive coding is a highly interesting candidate for the design of machine learning systems, e.g. due to the uncertainty-aware and hierarchically organized structure of predictive coding models.

Despite this, the influence of the predictive coding theory on machine learning and artificial intelligence, however, is less significant at the time of writing. There are many conceptual similarities between the state-of-the-art in machine learning, namely deep neural networks and the error back-propagation algorithm, and predictive coding models. Work that aims at directly connecting these fields, however, is still relatively sparse \cite{millidge2020predictive, millidge2021neural, rane2020prednet, ahmadi2019novel, oord2018representation}. Recently, it has been suggested that gradient based predictive coding, when it includes precision estimations, directly implements a form of Natural Gradient Descent, i.e. second order, uncertainty-aware optimization of the network’s parameters \cite{millidge2020predictive}. Within the domain of Deep Learning, Natural Gradient Descent (NGD) is a highly attractive optimization method, due to the speed and stability of convergence \cite{rattray1998natural, kunstner2019limitations}. In practice, however, NGD is oftentimes too computationally expensive and substantial effort has been devoted to methods that approximate the underlying second order statistics \cite{yao2020adahessian, zeiler2012adadelta}.

Still, despite these interesting connections between predictive coding and uncertainty-aware optimization of artificial neural networks, there is a lack of models that actually implement precision weighting in hierarchical models that scale to complex spatial and temporal inputs, such as they are common in deep learning.

Here we will address this lack of precision-aware predictive coding models and will try to clarify core conceptual questions that are still open with respect to top-down versus bottom-up processing with precision. Our method offers an attractive approach to uncertainty-aware optimization due to its distributed nature that is still directly compatible with the current methodology of training deep neural networks with (global) error backpropagation.

\section{Related work}

Existing work on predictive coding models in the domain of machine learning, and deep learning in particular, can roughly be divided into probabilistic, i.e. with explicit modeling of distributions or explicit sampling strategies, and deterministic models, without a stochastic part. Models implementing deterministic variants typically implement only selected concepts of the predictive coding theory, such as hierarchical error prediction or predictive representations for efficient spatio-temporal representation learning \cite{han2018deep, lotter2016deep, rane2020prednet, oord2018representation}. Predictive coding models with probabilistic structure have found application in the domains of robotics, sensory and brain signal prediction \cite{ofner2020balancing, ahmadi2019novel}. Other models explicitly compare deterministic and probabilistic implementations to predictive coding \cite{ahmadi2017bridging}. Gradient based predictive coding, the approach we will use here, is largely biologically plausible as all optimization can be described with Hebbian updates rules \cite{millidge2021neural,bogacz2017tutorial}. Gradient based predictive coding is deterministic based on the nature of its updates without explicit sampling. It does, however, allow to estimate the uncertainty of the prediction errors that drive the optimization of activities and weights in the network. It's these uncertainties that we will focus on in this work. Previous research revolving around gradient based predictive coding has illuminated its close connection to the error backpropagation algorithm, whose gradients have been shown to be directly approximated by discriminative predictive coding \cite{millidge2020predictive}. Predictive coding models have also been compared to other popular inference methods, such as the Kalman filter or variational inference in general \cite{millidge2021predictivereview}. While Kalman Filters have already been shown to implement a form of Natural Gradient Descent, the connection between natural gradients to predictive coding networks is yet to be explored in detail \cite{ollivier2018online}.

Second order optimization for deep neural networks is typically addressed by systems that approximate the second order statistics of the model with simpler and less computationally expensive approaches. Adam and RMSProp approximate the second order derivatives by computing the variance of the gradient \cite{kingma2014adam}. Other approaches compute approximate or exact second order derivatives directly, e.g. \cite{zeiler2012adadelta} and \cite{yao2020adahessian}.

\section{Precision and Natural Gradient Descent}

Estimating model parameters from noisy data samples is challenging. Since predictive coding networks possess multiple layers operating in parallel, each layer needs to be able to quantify the expected variance (or inverse precision) of neighboring activities. In predictive coding, this task is solved via the propagation of precision-weighted prediction errors. These weighted prediction errors measure the difference between the expected and observed precision of the measured activity. As we will see later, these precision-weighted prediction errors have a direct connection to the Fisher information, a central quantity in Natural Gradient Descent.

In information geometry, the quality of a parameter estimation rises and falls proportionally with the variance, or sharpness of the underlying probability density function (PDF). We refer to the gradient of the log-likelihood function $L$ as its score:

\begin{equation}
    u(\theta)=\frac{\partial}{\partial \theta}[\ln L(\theta)]
\end{equation}

This gradient, or first derivative, is zero at maximum and minimum points of the curve and can be used to find maximum likelihood estimates of the network's parameters $\theta$:

\begin{equation}
u(\hat{\theta})=0
\end{equation}

Under the regularity condition that the expectation of the score at the maximum likelihood (ML) estimate is zero, the Fisher Information $I(\theta)$ measures the variance of the score and thus the sharpness (or second derivative) of the log likelihood function:

\begin{equation}
I(\theta) =-\operatorname{var}[u(\theta)] =-E\left[\frac{\partial^{2} \ln [L(\theta)]}{\partial \theta^{2}}\right]
\end{equation}

A log likelihood function with a low amount of variance will have a sharp curve and high Fisher information, while high variance is reflected in low Fisher information. This means that, when we have access to the Fisher information of the log likelihood function, we can assess the accuracy of our estimation with respect to the function's parameters.

This property of the Fisher Information is used in Natural Gradient Descent as an adaptive learning rate during parameter updating. In practice, however, Natural Gradient Descent is computationally feasible only for models with relatively low complexity, since it's generally infeasible to compute and store the matrix for large parameter spaces. Furthermore, a direct computation of the Fisher information in the layers of a neural network is deemed not biologically plausible. 

More intuitively speaking, we can characterize precision as a measure of the repeatability of a measurement. If the network makes precise measurements of a constant quantity, the inferred value will be the same between repetitions and the likelihood function will be sharp with variance close to zero. Precision decreases with increasing amount of noise, either in the measured quantity or inherent in the measuring sensor (c.f. input noise and parameter dropout noise in deep neural networks).

\begin{figure}[h]
    \centering
    \includegraphics[width=0.4\textwidth]{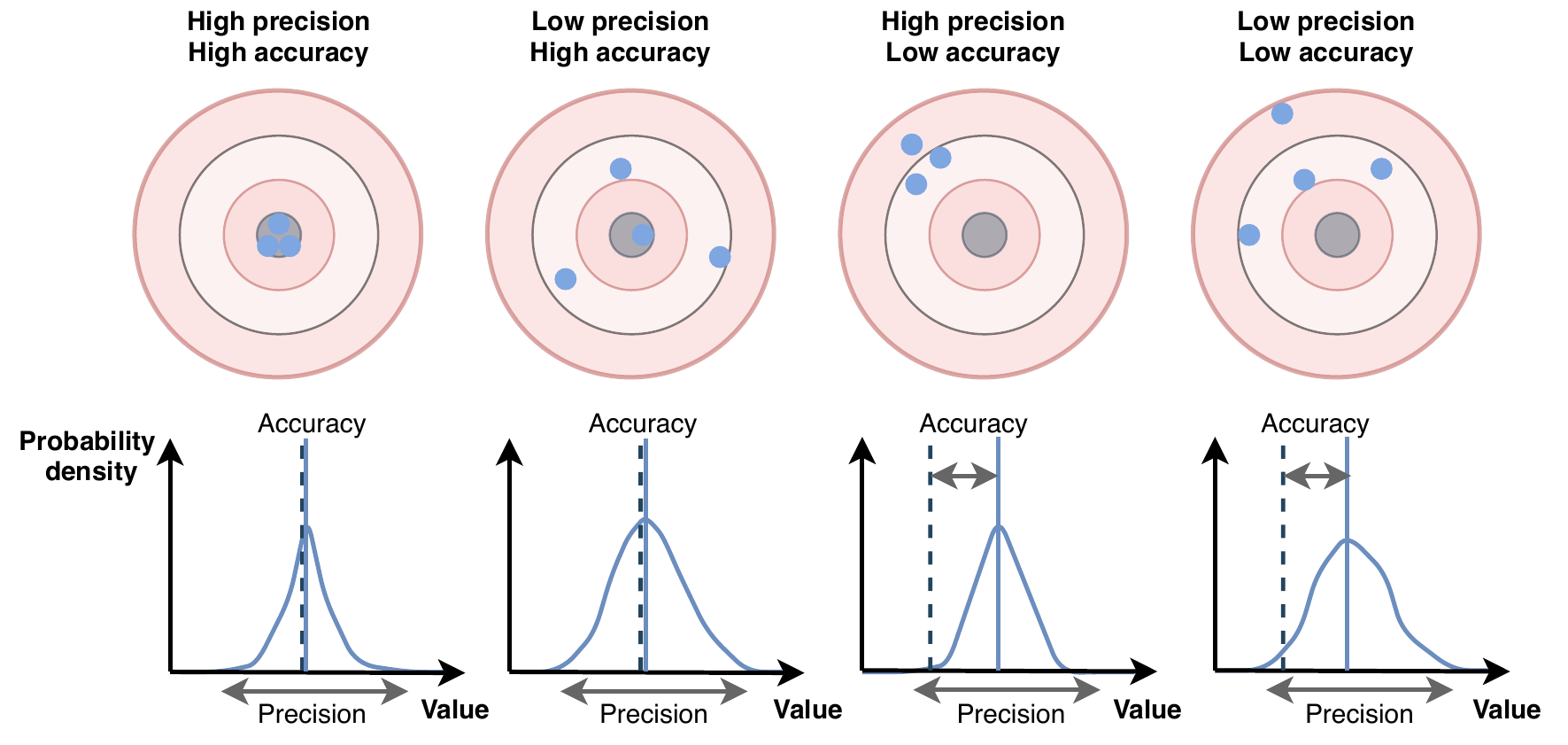}
    \caption{Precision and accuracy on a simple toy example.}
    \label{fig:precacc}
\end{figure}

\section{Method}

Our suggested predictive coding network is implemented via stacked and only locally connected network modules that are trained via standard gradient descent and backpropagation of the relevant prediction errors. This allows to effortlessly merge exact inference within the modules via backpropagation (through potentially very deep and nonlinear functions) to the distributed, uncertainty aware and bi-directional inference of the PC network. When each module contains a single layer, the network has a structure that is equivalent to the models discussed in \cite{bogacz2017tutorial} and \cite{millidge2021predictivereview}, i.e. simplifies to gradient based predictive coding. 

\begin{figure}[h]
    \centering
    \includegraphics[width=0.5\textwidth]{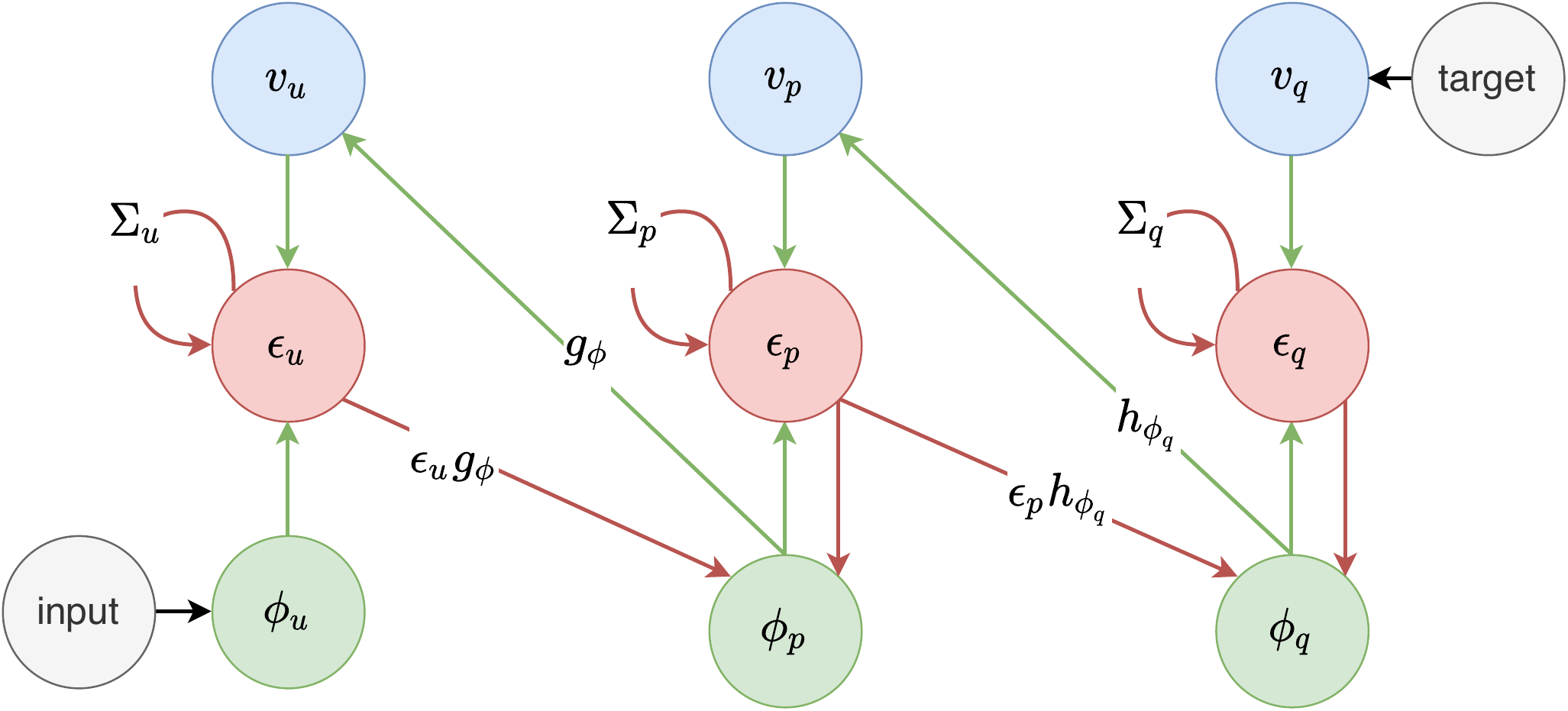}
    \caption{Predictive coding model with three layers.}
    \label{fig:precacc}
\end{figure}

However, in contrast to these networks, our network actually learns the precision embedded in its hierarchical structure and does not simply set them to a constant value. We refer to this sort of implementation as "local backpropagation" and hope that it accelerates the integration of concepts from predictive coding into the established methodology of training deep neural networks. Figure \ref{fig:precacc} shows an overview of the predictive coding network with precision estimation, that is employed in this work and will be discussed in detail in later sections.

\subsection{Precision estimation from local prediction errors}

\begin{figure}[h]
\begin{centering}
\includegraphics[width=0.3\textwidth]{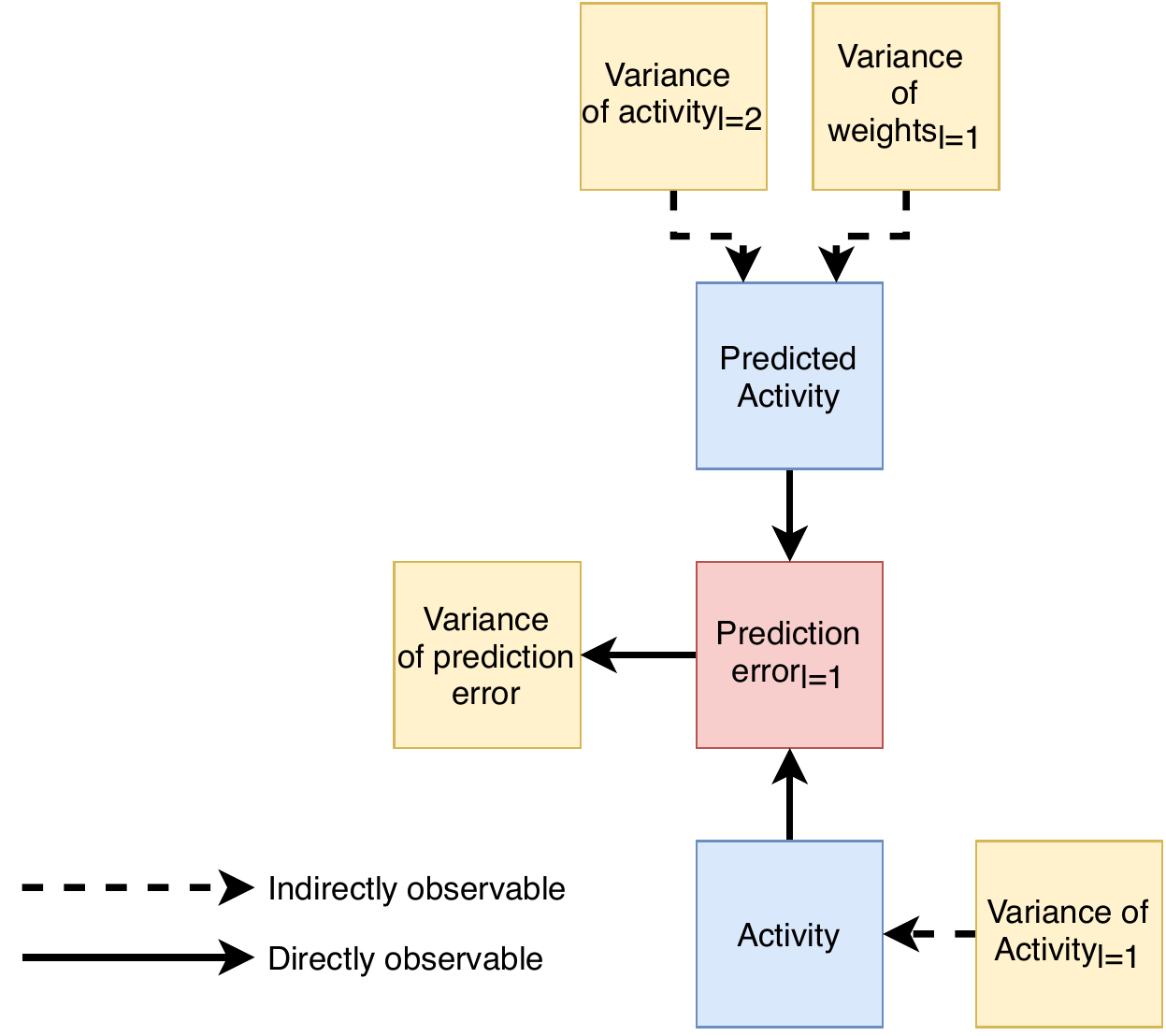}
\caption{Directly and indirectly observable sources of uncertainty in gradient based predictive coding.}
\label{fig:sources_of_uncertainty}
\end{centering}
\end{figure}

Figure \ref{fig:sources_of_uncertainty} shows an overview of the sources of uncertainty that arise when estimating the precision of observed prediction errors in each layer of the network. When all inherent timescales in the network are the same, i.e. all learned quantities are updated at the same frequency, then only the prediction error is directly observable. However, with increasing separation of timescales (e.g. weights update slower than activities), the remaining sources of uncertainty can be observed more directly. While some choice of timescales is inherent to all neural network models, here we specifically look at their dynamic modulation.

These timescales can either be included explicitly into the model (e.g. via learning rate priors or fixed amount of update steps for each variable). The timescales can also be encoded implicitly, e.g. through the adaptation of learning rates (e.g. high learning rate is similar to making more updates). This hints at the aspect of learned (dynamic) timescales in precision weighted predictive coding. For example, the estimated precision in a model with rapid weight updates looks vastly different to a model with slow weights learning rate. Since the learning rate is directly dependent on the estimated precision, this results in a feedback loop between currently inferred states, network parameters and learning rates.

Following the notation of \cite{millidge2021predictivereview} and \cite{millidge2020predictive}, the Free Energy can be defined as:
\begin{equation}
\mathcal{F}=\sum_{l=1}^{L} \Sigma_{l}^{-1} \epsilon_{l}^{2}+\ln 2 \pi \Sigma_{l}^{-1}
\end{equation}

with prediction errors $\epsilon_{l}$ between mean activity $\mu_{l}$ and predicted mean activity $\hat{\mu}_{l}$ at layer $l$:
\begin{equation}
\epsilon_{l}=\mu_{l}-f_{l}\left(\theta_{l+1}, \mu_{l+1}\right)
\end{equation}

The dynamics of the mean activity $\mu_{l}$ and synaptic weights $\theta_{l}$ in layer l compute as:

\begin{equation}
\begin{aligned}
\frac{d \mu}{d t} =-\frac{\partial \mathcal{F}}{\partial \mu}=\Sigma_{l+1}^{-1} \epsilon_{o} \frac{\partial f}{\partial \mu} \theta^{T}-\Sigma_{l}^{-1} \epsilon_{x} \\
\frac{d \theta_{l}}{d t} =\frac{\partial \mathcal{F}}{\partial \theta_{l}}=-\Sigma_{l+1}^{-1} \epsilon_{o} \frac{\partial f}{\partial \theta_{l}} \mu^{T} \\
\frac{d \theta_{l+1}}{d t} =\frac{\partial \mathcal{F}}{\partial \theta_{l+1}}=-\Sigma_{l}^{-1} \epsilon_{x} \frac{\partial g}{\partial \theta_{l+1}} \bar{\mu}^{T}
\end{aligned}
\end{equation}

with  \begin{equation}
\epsilon_{o}=o-f\left(\mu, \theta_{l}\right) \text { and } \epsilon_{x}=\mu-g\left(\bar{\mu}, \theta_{l+1}\right)
\end{equation}

The layer's precision $\Sigma_{l}^{-1}$ of layer $i$ is the inverse variance, which is computed as:
\begin{equation}
\begin{aligned}
\frac{d \Sigma}{d t}=-\frac{\partial \mathcal{F}}{\partial \Sigma} =\Sigma_{l}^{-T} \epsilon_{l} \epsilon_{l}^{T} \Sigma_{l}^{-T}-\Sigma_{l} \\
=\Sigma_{l}^{-1} \epsilon_{l} \epsilon_{l}^{T} \Sigma_{l}^{-T}-\Sigma_{l} \\
=\tilde{\epsilon}_{l} \tilde{\epsilon}_{l}^{T}-\Sigma_{l}
\end{aligned}
\end{equation}

where $\tilde{\epsilon}_{l}$ are the precision weighted prediction errors:
\begin{equation}
\tilde{\epsilon}_{l}=\sum_{l}^{-1} \epsilon_{l}
\end{equation}

Update rules for the layer's activities $v_{i}$ and weights $\theta_{i}$:

\begin{equation} 
v^{t+1} =v^{t}+\eta \Sigma^{-1} \frac{\partial \tilde{\mathcal{F}}}{\partial v_{i}}
\end{equation}
\begin{equation} 
{\theta^{t+1} =\theta^{t}+\eta \Sigma^{-1} \frac{\partial \tilde{\mathcal{F}}}{\partial \theta_{i}}} \end{equation}

The Fisher information of the layer's activity is equal to the estimated precision error precision at that layer \cite{millidge2021predictivereview}:

\begin{equation}
\begin{aligned}
\mathcal{G}\left(\mathcal{F}, \mu_{l}\right) &=\mathbb{E}\left[\frac{\partial^{2}}{\partial \mu_{l}^{2}} \mathcal{F}\right] \\
&=\mathbb{E}\left[\frac{\partial^{2}}{\partial \mu_{l}^{2}}\left(\left(\mu_{l}-f\left(\theta_{l} \mu_{l+1}\right)\right)^{T} \Sigma_{l}\left(\mu_{l}-f\left(\theta_{l} \mu_{l+1}\right)\right)\right)\right] \\
&=\mathbb{E}\left[\frac{\partial^{2}}{\partial \mu_{l}^{2}}\left(\mu_{l} \Sigma^{-1} \mu_{l}\right)\right] \\
&=\mathbb{E}\left[\Sigma_{l}^{-1}\right] \\
&=\Sigma^{-1}
\end{aligned}
\end{equation}

Similarly, the Fisher information with respect to the weights can be expressed as the layer's estimated precision of prediction error, scaled by the variance of activities at the next higher layer. A slightly longer derivation of this connection can be found in appendix A. 

{
\begin{equation}
\begin{aligned}
\mathcal{G}\left(\mathcal{F}, \theta_{l}\right) &=\mathbb{E}\left[\frac{\partial^{2}}{\partial \theta_{l}^{2}} \mathcal{F}\right] \\
&=\mathbb{E}\left[\frac{\partial^{2}}{\partial \theta_{l}^{2}}\left(\left(\mu_{l}-f\left(\theta_{l} \mu_{l+1}\right)\right) \Sigma_{l}\left(\mu_{l}-f\left(\theta_{l} \mu_{l+1}\right)\right)\right)\right] \\
&\left.=\mathbb{E}\left[\frac{\partial^{2}}{\partial \theta_{l}^{2}}\left(\theta_{l} \mu_{l+1}\right)^{T} \Sigma_{l} \theta_{l} \mu_{l+1}\right)\right] \\
&=\Sigma_{l}^{-1} \mathbb{E}\left[\mu_{l+1} \mu_{l+1}^{T}\right] \\
&=\Sigma_{l}^{-1} \mathbb{V}\left[\mu_{l+1}\right]
\end{aligned}
\end{equation}
}

\subsection{Precision estimation with fixed predictions or inputs}

The exact role of bottom-up and top-down propagation of precision is a largely under-explored topic in predictive coding. The often mentioned aspect of precision propagation within PC networks stands in contrast to the actual precision updates that are intrinsic to each layer, without the need to explicitly exchange information about expected precision between layers. We hope that our model clarifies some of these issues. The inferred variance in each layer depends on the error on two constantly changing input sources: The posterior activities inferred from a stream of bottom-up prediction errors and the top-down predictions from the next higher layer.

In a model with fixed weights and posterior activities in hidden layers one can iteratively sample multiple predictions of the input per data point, or sample multiple observations of the same data point (e.g. the same input image changed by additive noise). We can then update the variance estimate until convergence and only then progress to the next activity update. Comparison of such different types of inference on variance with frozen model weights is discussed in section \ref{variance_estimation}. In contrast, when model parameters and activities are updated during variance estimation, the precision-weighted prediction errors naturally scale the learning rates of updates for each variable individually. Experiments conducted with these models are presented in sections \ref{variance_learning_noisy} and \label{weights_learning}. Since the learning rates directly affect the magnitude of change over time for each parameter, this resembles implicitly learned timescales in predictive coding models.

\section{Results}
\label{results}

\subsection{Variance estimation}
\label{variance_estimation}

\subsubsection{Unsupervised Predictive Coding}
Experimental setup:
\begin{enumerate}
    \item Images are auto-encoded by repeatedly running forward and backward passes through the entire network. Still, information is updated only locally between layers.
    \item We can compare the inference process to an autoencoder, where the activity units of the deepest layer play the role of the latent embedding. Decoding in autoencoders can be compared to the backward prediction pass in predictive coding. Similarly, the encoding process can be equated to the forward propagation of prediction errors.
    \item We evaluate the unsupervised model with respect to reconstruction errors and the level of disentanglement in the deepest layer.
    \item The PC network resembles a hierarchical autoencoder, since we can decode the activities (i.e. latent states) in any of the hidden layers, too. We expect the information in these layers to be less disentangled.
\end{enumerate}

\begin{figure}[h]
\begin{subfigure}[t]{0.5\textwidth}
         \centering
         \includegraphics[width=1\textwidth]{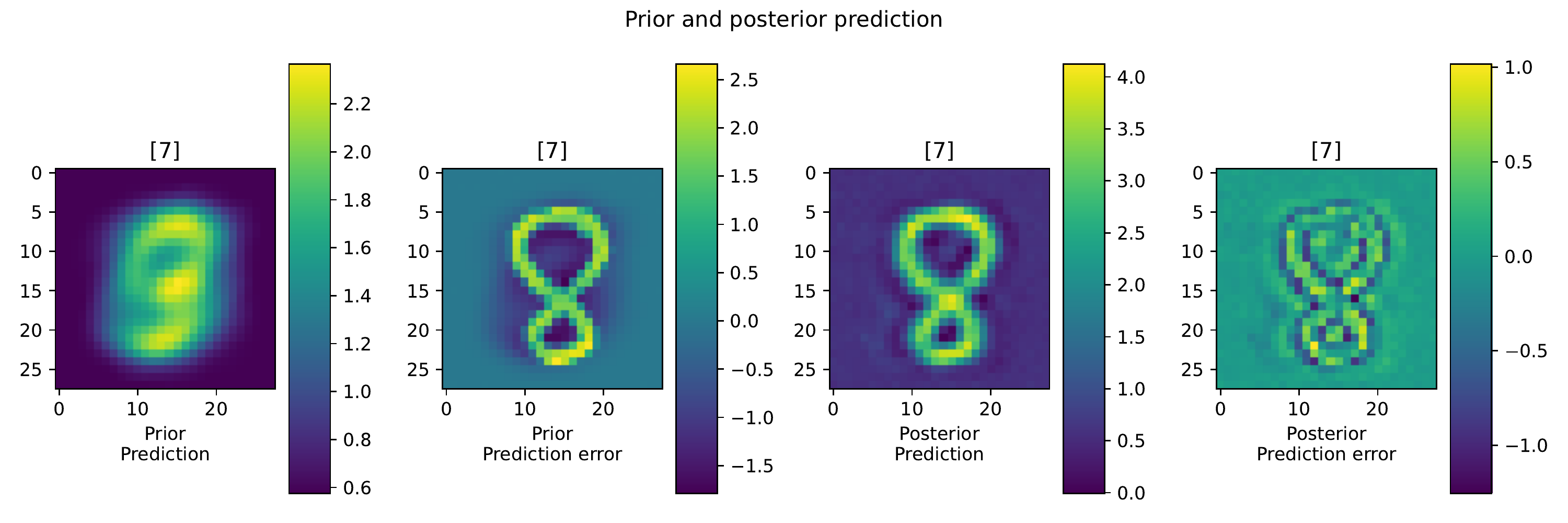}
         \caption{Autoencoded digits of the MNIST dataset.}
         \label{fig:autoencoded-digits-MNIST}
     \end{subfigure}
     \hfill
     \begin{subfigure}[t]{0.5\textwidth}
         \centering
	    \includegraphics[width=0.4\textwidth]{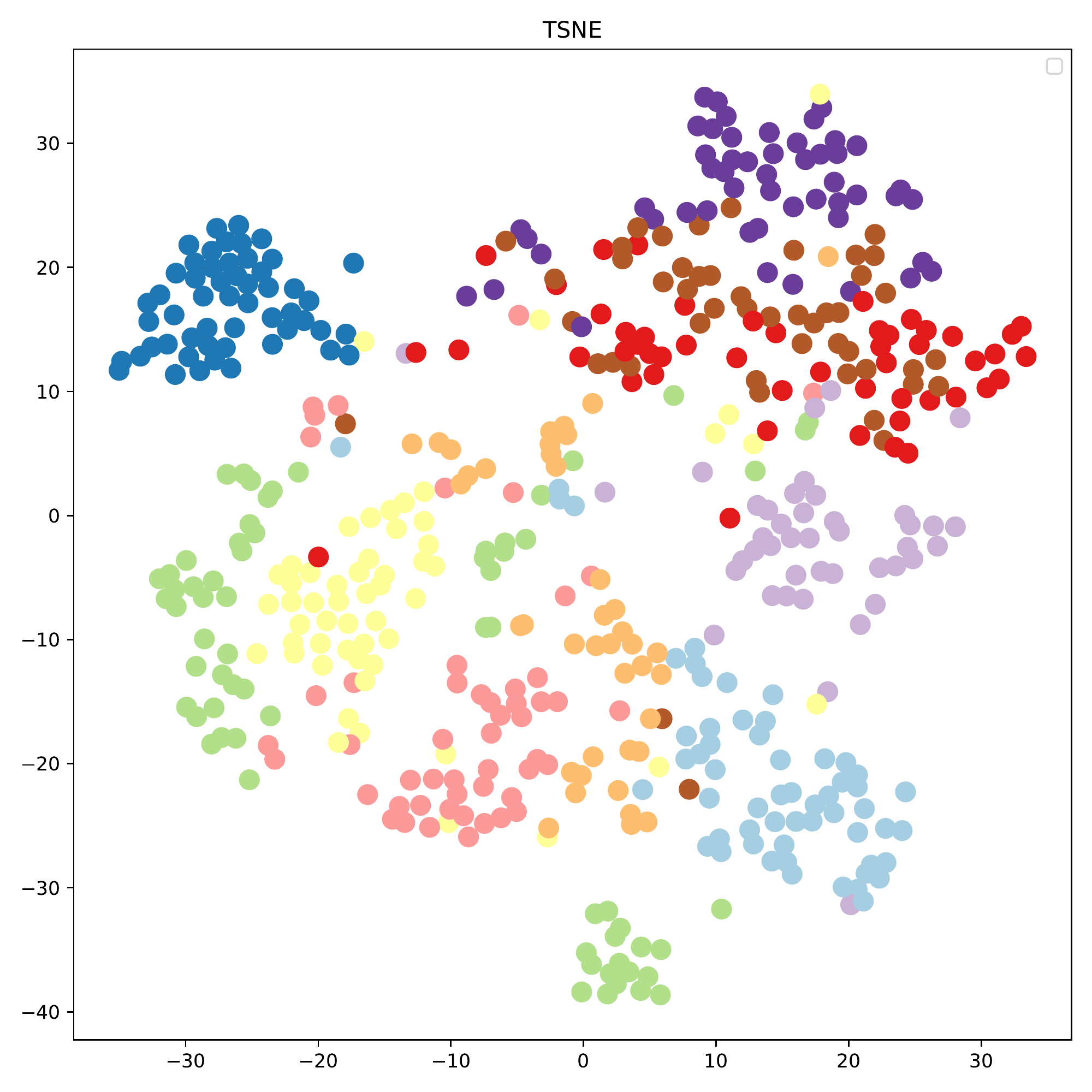}
         \caption{Learned embedding visualization (t-SNE) on the MNIST test set.}
         \label{fig:embedding_visualization_MNIST}
     \end{subfigure}
     \caption{Visualized reconstructions and embeddings after training for 10 epochs on the MNIST dataset.}
\end{figure}

Despite there being no explicit sampling step, the embeddings for MNIST show a clear disentanglement when analyzed with t-SNE, hinting at the close connections between predictive coding and variational inference.

\subsubsection{Variance estimation with noisy data and weights}
\label{variance_learning_noisy}

Experimental setup:
\begin{enumerate}
    \item controlled noise on data with mean zero and variance of ten
    \item controlled noise on weights using dropout at test time (MC dropout)
    \item uncertainty is updated in all layers in parallel, only the input layer is evaluated
\end{enumerate}

\begin{figure}[H]

  \subfloat[Variance estimation with fixed activity posterior.]{
	   \centering
	   \includegraphics[width=0.48\textwidth]{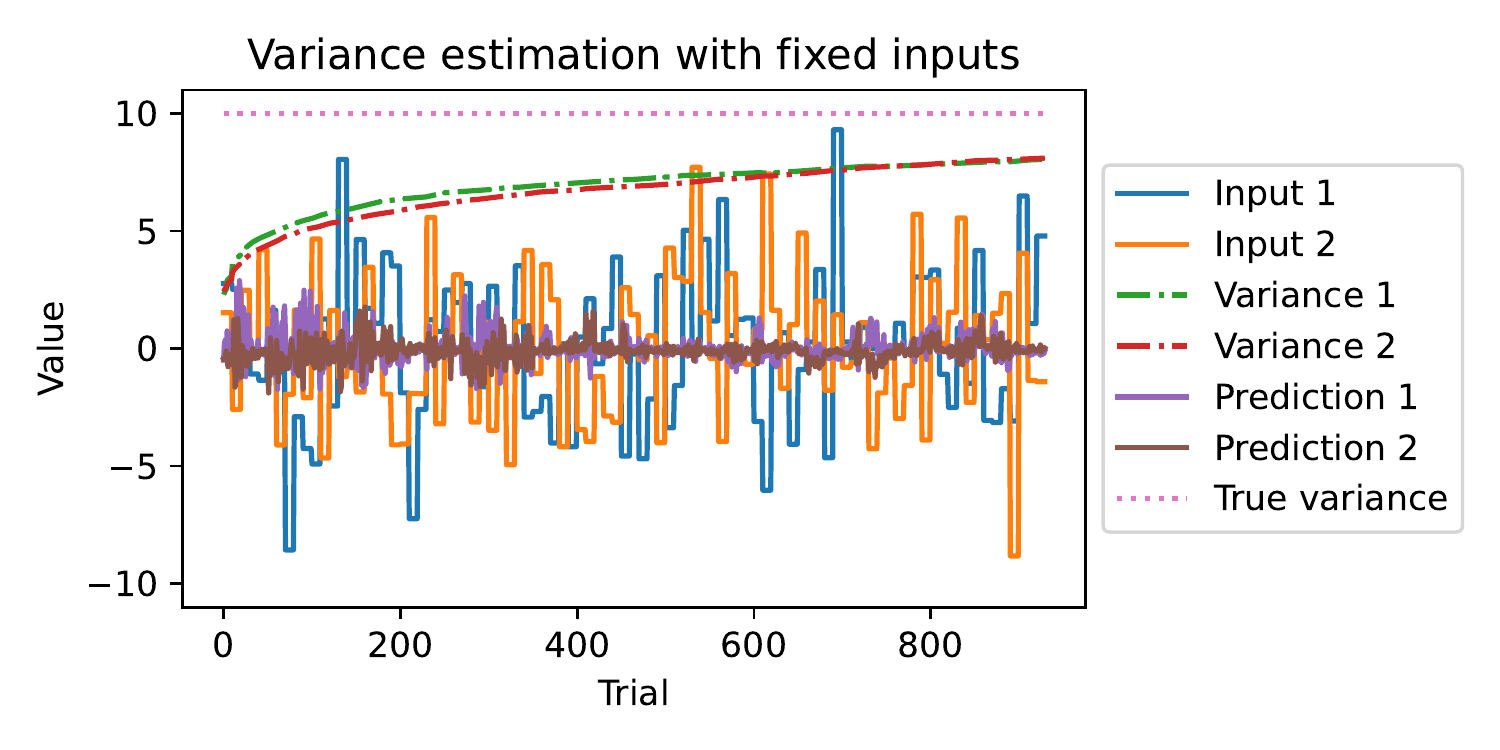}}
  \subfloat[Variance estimation with fixed top-down predictions.]{
	   \includegraphics[width=0.48\textwidth]{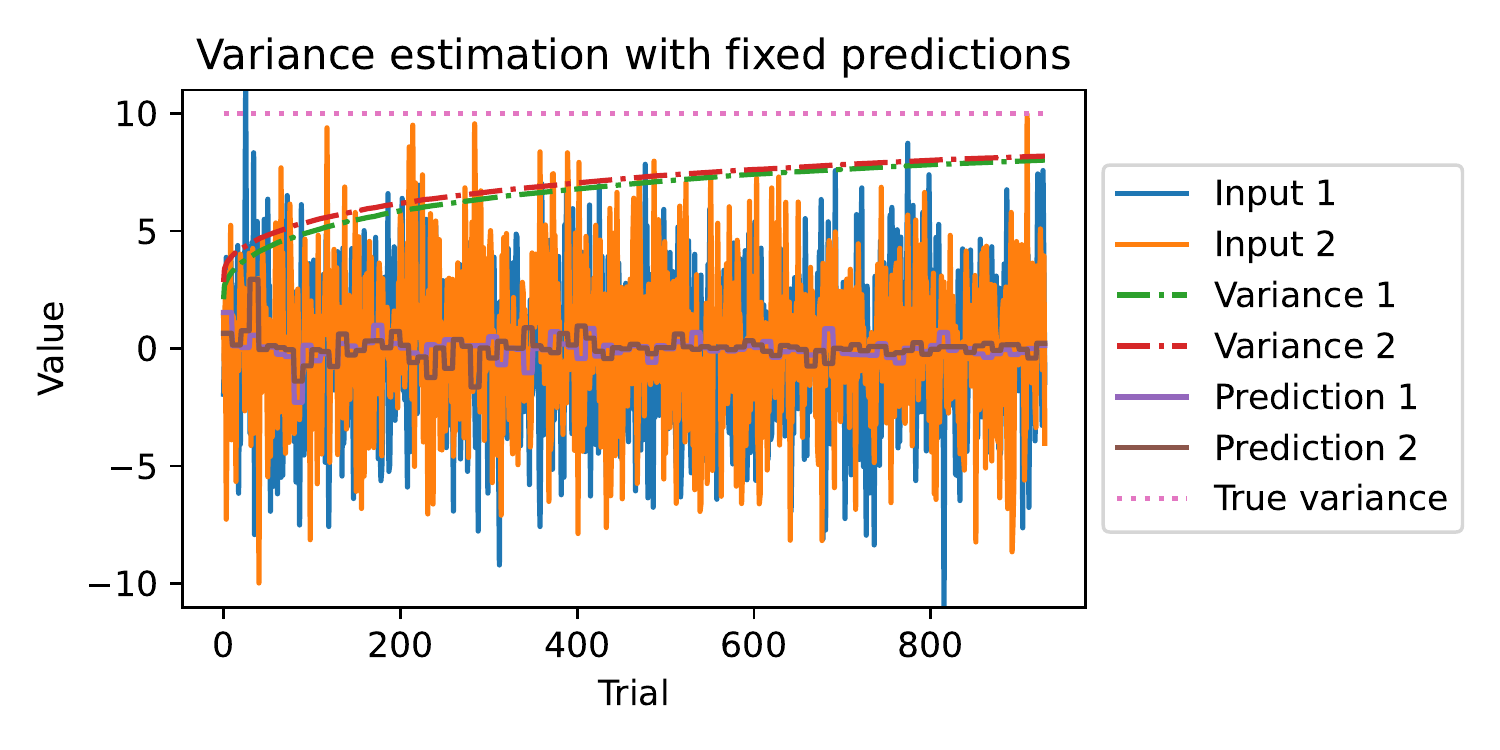}}
\caption{Estimating variance during the prediction of constant inputs with zero mean and additive Gaussian noise with variance 10. Shown are inferred variance values with respect to either fixed inputs (a) or predictions (b) during variance updating. Both variants successfully infer the observed variance, despite the noisiness (30\% dropout) of the top-down prediction.}
\end{figure}

For fixed inputs or predictions, the inferred variance values converge quickly: Variance learning rate 0.1, 10-100 updates. Variance estimates without fixed inputs or predictions still infer the true data noise for inputs with constant mean.

\subsection{Weights learning with and without precision weighting}
\label{weights_learning}

Experimental setup:
\begin{enumerate}
    \item single layer PC model with activity set to be the input and top-down prediction computed via targets and weights
    \item the network is trained to infer the precision of prediction error and the correct weights at the same time
    \item single unit input: controlled noise on data with mean 5 and variance of 2
    \item we test two conditions: controlled noise on target (mean 2,5 and variance 2) and without target noise 
    \item when correctly learned, the inferred weight value should be 2 and the variance should be 2 without target noise and 4 with target noise
\end{enumerate}

The learned precision of prediction error accounts for uncertainty from (bottom-up) activity and (top-down) prediction. Note that the (top-down) prediction splits up into fluctuations caused by the activity at the higher layer and fluctuations from the weights updates.

\begin{figure}[H]
	   \centering
	   \includegraphics[width=0.4\textwidth]{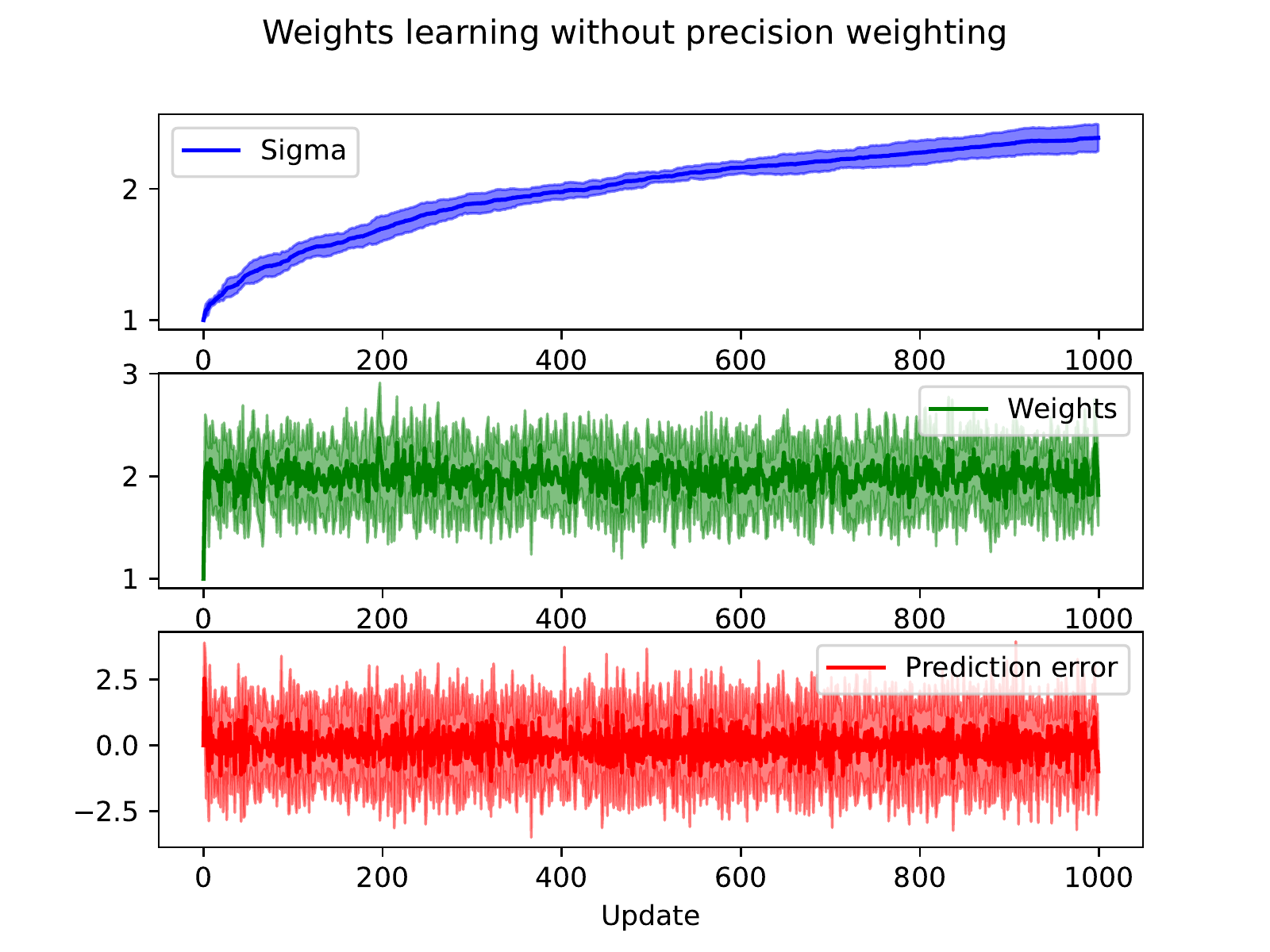}
	   \includegraphics[width=0.4\textwidth]{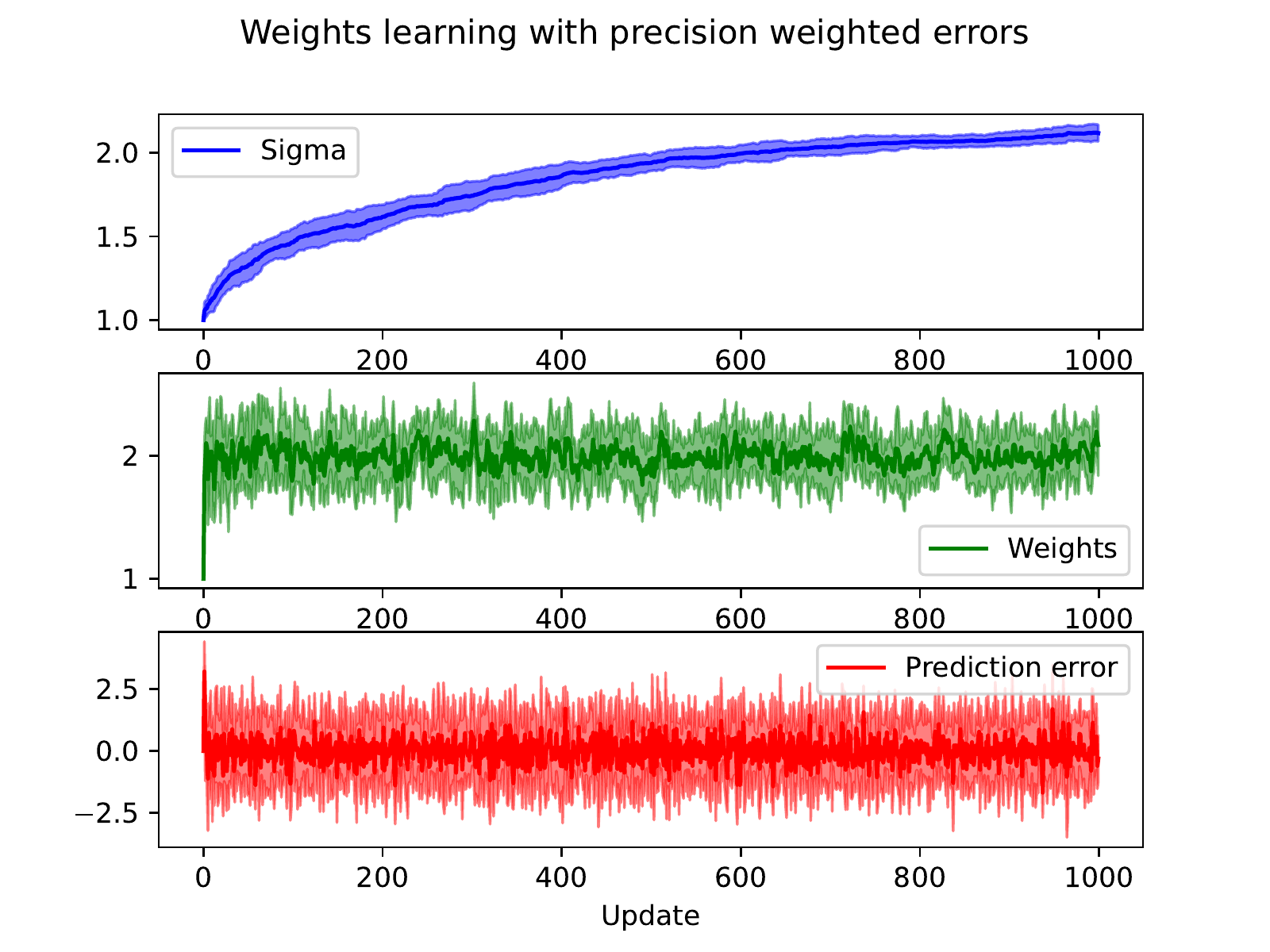}
\caption{Weights learning with and without precision weighted prediction errors. Observed inputs have a constant variance of 2. The network trained with precision weighted errors more accurately infers the observed variance.}
\end{figure}

\begin{figure}[H]
	   \centering
	   \includegraphics[width=0.4\textwidth]{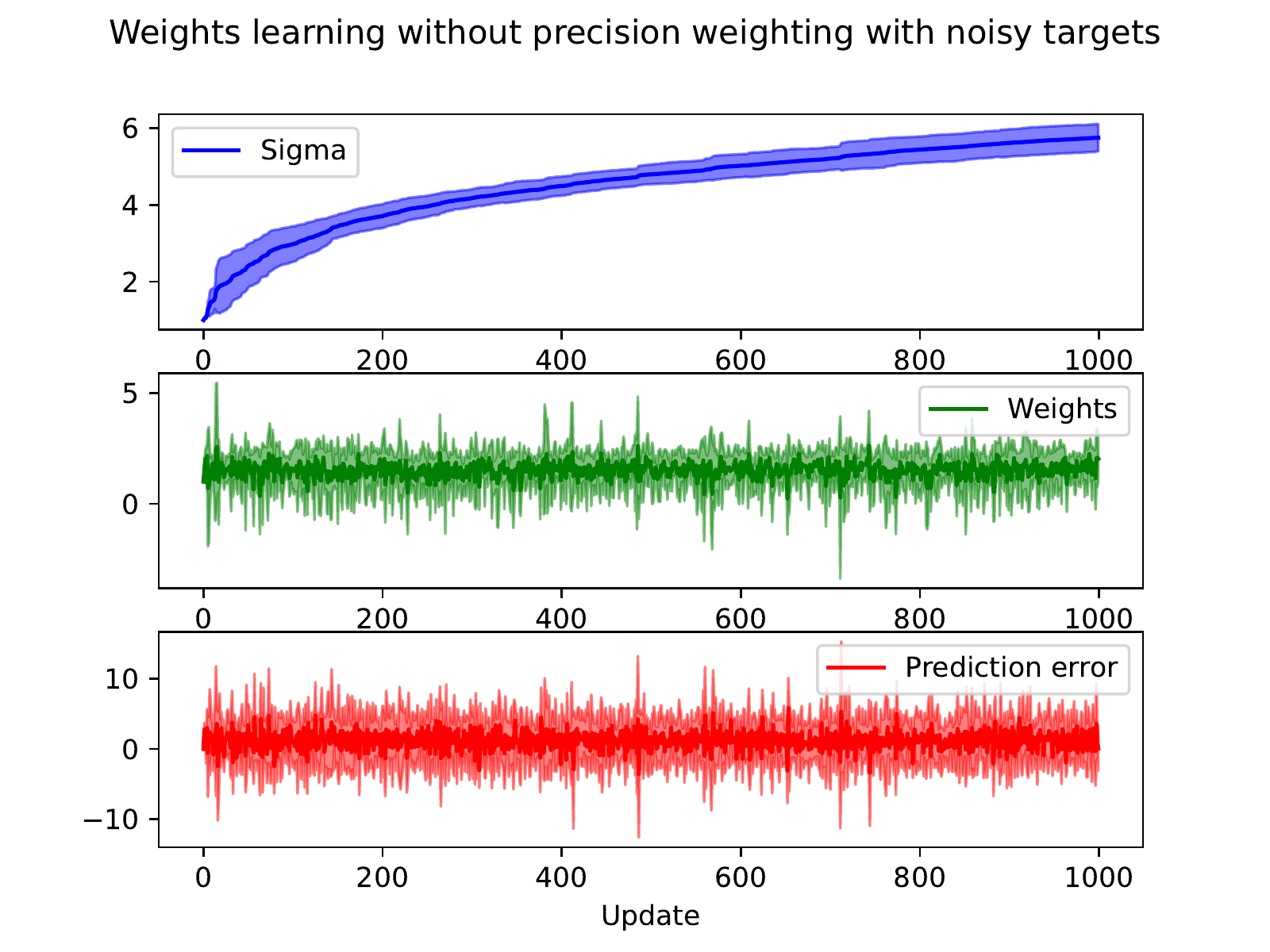}
	   \includegraphics[width=0.4\textwidth]{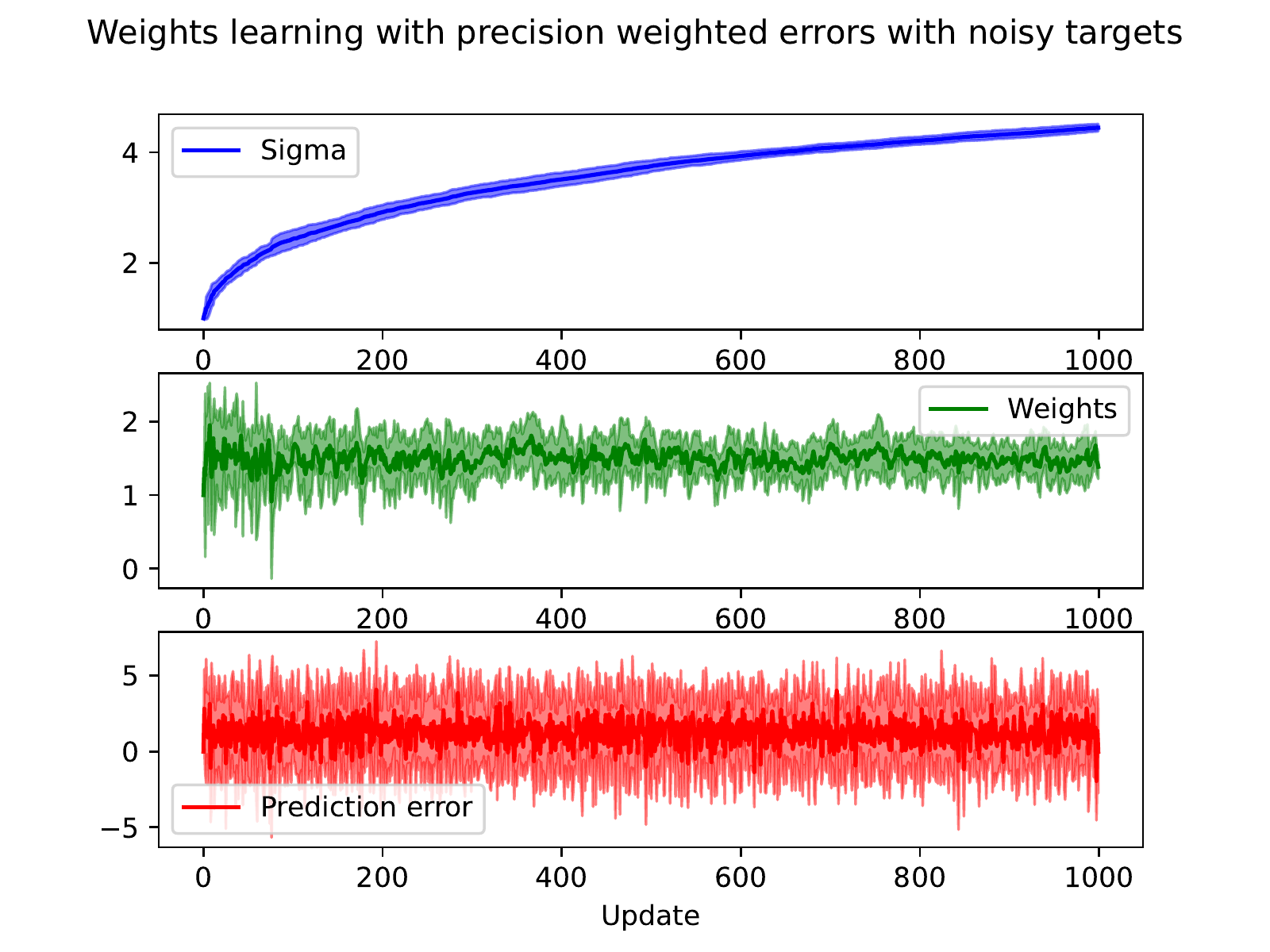}
\caption{Weights learning with and without precision weighted prediction errors. Observed inputs and targets have a constant variance of 2. In this setting, the learned precision needs to integrate observed variances from two sources: the bottom-up inputs (from the noisy data) and the top-down predicted input (from the noisy target). Again, only the weights updates that include precision weighting infer the correct total variance value of 4. During training without precision weighting, the standard deviation of both inferred weights and precision values is significantly higher.}
\end{figure}

\subsection{Noisy MNIST dataset}
\subsubsection{MNIST classification}
\label{MNISTclassification}

\begin{center}
\begin{tabular}{||c c c c c c||} 
 \hline
Model & Input noise & Target noise & Hidden weights & Prediction Dropout & Accuracy\\ [0.5ex] 
 \hline\hline
PC-SGD & 0. & 0. & 256 & 0.3 & 0.90 \\ 
PC-Adadelta & 0. & 0. & 256 & 0.3 & 0.93 \\ 
 \hline
\end{tabular}
\end{center}

\subsubsection{MNIST reconstruction}
\label{MNISTreconstruction}

\begin{figure}[H]
\begin{subfigure}[t]{0.48\textwidth}
    \includegraphics[width=\textwidth]{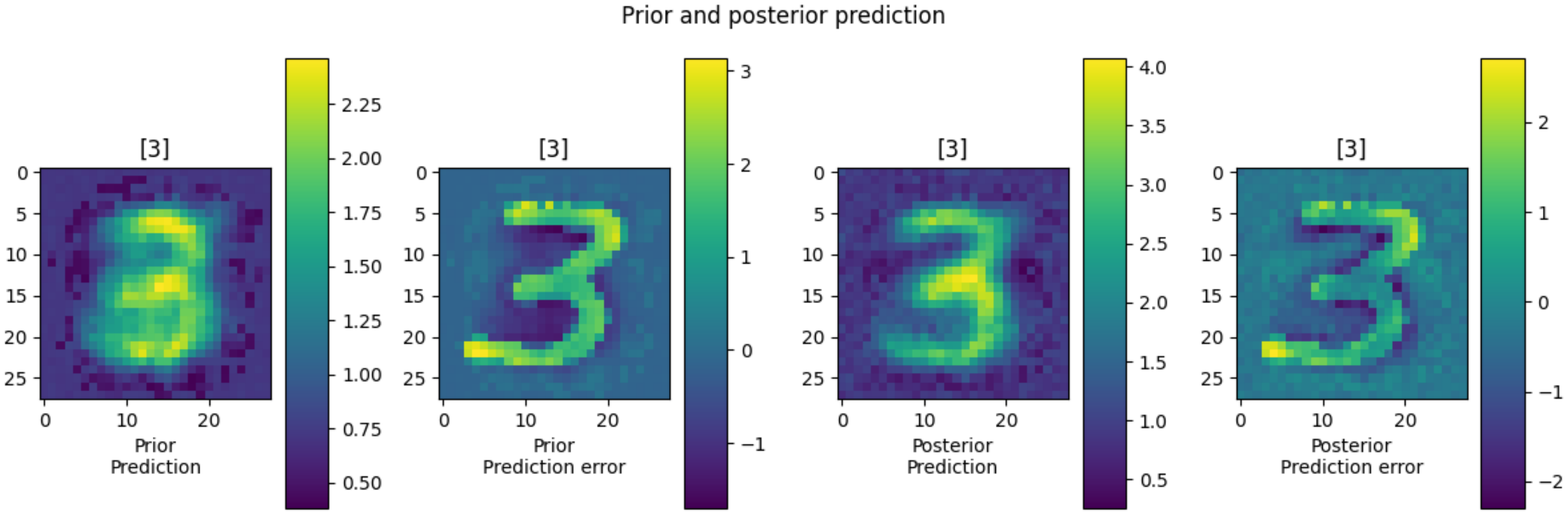}
    \subcaption{Comparison of prior predictions from the target label and posterior predictions after 10 updates of hidden activities using the prediction error on the true input.}
\end{subfigure}
\hfill
\begin{subfigure}[t]{0.48\textwidth}
         \centering
         \includegraphics[width=\textwidth]{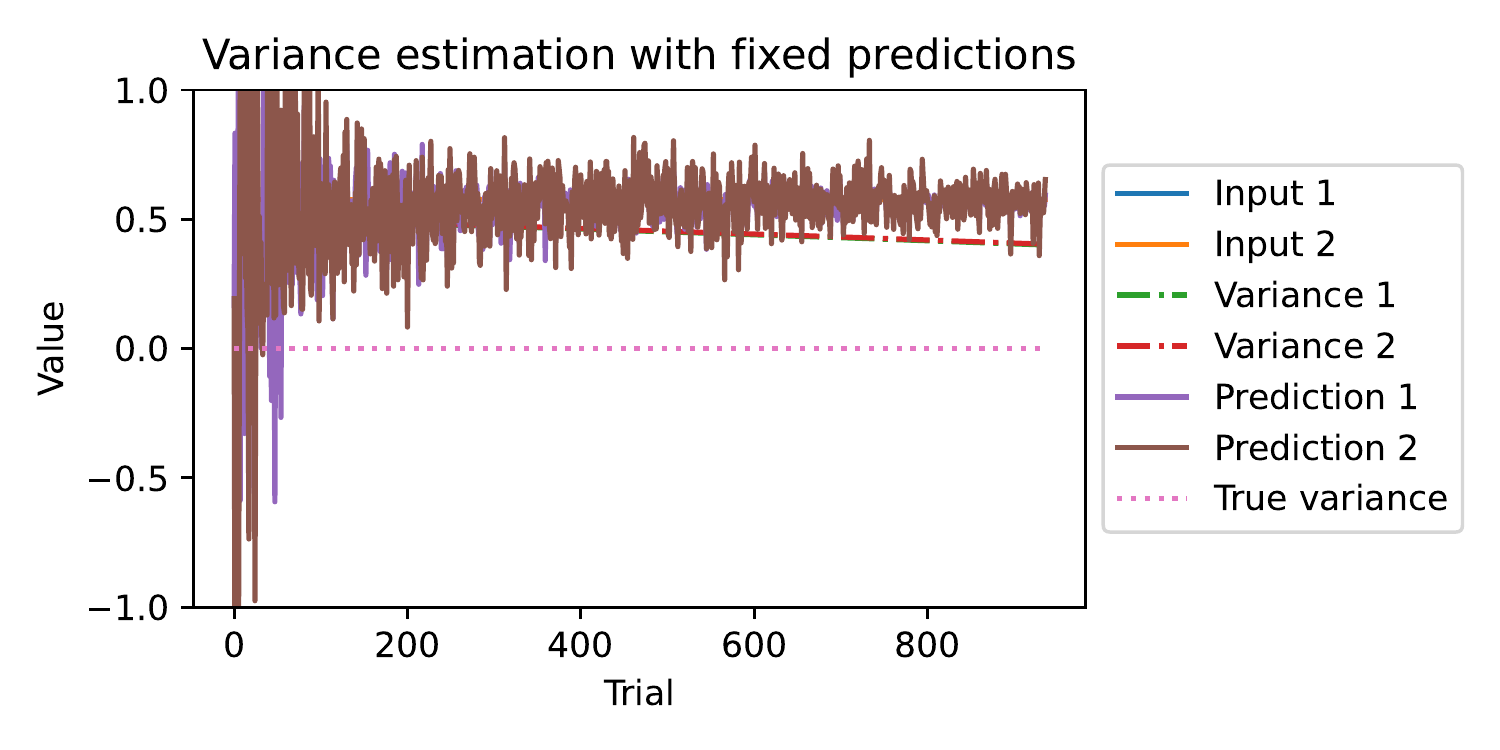}
         \caption{Estimated variance in the input layer. True variance is zero since no additive noise was added to the original MNIST dataset.}
         \label{fig:MNIST_completevariance}
     \end{subfigure}
\caption{Estimating variance during the prediction of constant inputs with zero mean and additive Gaussian noise with variance 10. Shown are inferred variance values with respect to either fixed inputs a) or predictions b) during variance updating. Both variants successfully infer the observed variance, despite the noisiness (30\% dropout) of the top-down prediction.}
\end{figure}

\printbibliography

\newpage 

\appendix
\section{Appendix A}

\subsection{Fisher information with respect to the Free Energy and model weights}

\begin{equation}
\begin{aligned}
\mathcal{F} =D_{\mathrm{KL}}[q(x \mid o ; \phi) \| p(o, x ; \theta)] \\
=\underbrace{\mathbb{E}_{q(x \mid o ; \phi)}[\ln q(x \mid o ; \phi)]}_{\text {Entropy }}-\underbrace{\mathbb{E}_{q(x \mid o ; \phi)}[\ln p(o, x ; \theta)]}_{\text {Energy }}\\
=-\underbrace{\mathbb{E}_{q(x \mid o ; \phi)}[\ln p(o, x ; \theta)]}_{\text {Energy }}
\end{aligned}
\end{equation}

Where the entropy of the dirac-delta distribution is 0 and only the energy term is non-zero.

The variational free energy is the sum of the prediction errors at each layer:
\begin{equation}
\mathcal{F}=\sum_{l=1}^{L} \Sigma_{l}^{-1} \epsilon_{l}^{2}+\ln 2 \pi \Sigma_{l}\\
\end{equation}

with errors $\epsilon_{l}=\mu_{l}-f_{l}\left(\theta_{l+1}, \mu_{l+1}\right)$.

The Fisher information with respect to the weights of the model is defined as the expected value of the second derivative of the variational free energy:

\begin{equation}
\begin{aligned}
(\sum_{l=1}^{L} \Sigma_{l}^{-1} \epsilon_{l}^{2}+ \ln 2 \pi \Sigma_{l}) &= \mathbb{E}\left[\frac{\partial^{2}}{\partial \theta_{l}^{2}} \mathcal{F}\right] \\
&= \mathbb{E} \left[ \frac{ \partial^{2}}{ \partial \theta_{l}^{2}}( \sum_{l=1}^{L} \Sigma_{l}^{-1} \epsilon_{l}^{2}+ \ln 2 \pi \Sigma_{l}) \right] \\
&=\mathbb{E}\left[\frac{\partial^{2}}{\partial \theta_{l}^{2}} (\sum_{l=1}^{L} \Sigma_{l}^{-1} \epsilon_{l}^{2})\right] \\
&=\mathbb{E}\left[\frac{\partial^{2}}{\partial \theta_{l}^{2}} \sum_{l=2}^{L} \Sigma_{l-1}^{-1} (f_{l-1}\left(\theta_{l}, \mu_{l}\right))^{2} \right] \\
&=\mathbb{E}\left[\frac{\partial^{2}}{\partial \theta_{l}^{2}} \sum_{l=2}^{L} \Sigma_{l-1}^{-1} f_{l-1}\left(\theta_{l}, \mu_{l}\right) f_{l-1}\left(\theta_{l}, \mu_{l}\right) \right] \\
&=\mathbb{E}\left[\frac{\partial^{2}}{\partial \theta_{l}^{2}} \sum_{l=2}^{L} \Sigma_{l-1}^{-1} (\theta_{l}\mu_{l}) (\theta_{l}\mu_{l}) \right] \\
&=\mathbb{E}\left[\frac{\partial^{2}}{\partial \theta_{l}^{2}} \sum_{l=2}^{L} \Sigma_{l-1}^{-1}\theta_{l}^{2}\mu_{l}^{2} \right] \\
&=\sum_{l=1}^{L} \Sigma_{l}^{-1} \mathbb{E} \left[\mu_{l+1}^{2} \right] \\
&=\sum_{l=1}^{L} \Sigma_{l}^{-1} \mathbb{V} \left[\mu_{l+1} \right] \\
\end{aligned}
\end{equation}

\end{document}